\documentclass[conference]{IEEEtran}
\pdfoutput=1
\IEEEoverridecommandlockouts
\usepackage{cite}
\usepackage{amsmath,amssymb,amsfonts}
\usepackage{algorithmic}
\usepackage{graphicx}
\usepackage{textcomp}
\usepackage{xcolor}
\usepackage{savesym}
\savesymbol{checkmark}
\usepackage{lipsum}
\usepackage[ruled,vlined]{algorithm2e}
\usepackage{enumerate}
\usepackage{tabularx,booktabs}
\usepackage{dingbat}
\usepackage{diagbox}
\usepackage[hidelinks]{hyperref}
\usepackage{multicol}
\usepackage{float}
\usepackage{color}
\usepackage{subcaption}
\usepackage{multirow}

\restylefloat{table}
\renewcommand{\figurename}{Figure}
\usepackage{fancyhdr} 
\def\BibTeX{{\rm B\kern-.05em{\sc i\kern-.025em b}\kern-.08em
    T\kern-.1667em\lower.7ex\hbox{E}\kern-.125emX}}

\begin{document}
\pagestyle{fancy}

\title{DSBERT:Unsupervised Dialogue Structure learning with BERT}

\author{
\IEEEauthorblockN{
Bingkun Chen\IEEEauthorrefmark{2},
Shaobing Dai\thanks{* Shaobing Dai is the corresponding author.}\IEEEauthorrefmark{2}\IEEEauthorrefmark{1},
Yang Li\IEEEauthorrefmark{2},
Shenghua Zheng\IEEEauthorrefmark{2}, and
Lei Liao \IEEEauthorrefmark{2}
}
\IEEEauthorblockA{
\IEEEauthorrefmark{2}QingNiuZhiSheng Technology Company, China, \href{mailto:xxx@xxx,xxx@xxx}{\{chenbk, daisb, liyang, zhengsh, liaolei\}@qnzsai.com}
}
}


\maketitle
\begin{abstract}

Unsupervised dialogue structure learning is an important and meaningful task in natural language processing. The extracted dialogue structure and process can help analyze human dialogue, and play a vital role in the design and evaluation of dialogue systems. The traditional dialogue system requires experts to manually design the dialogue structure, which is very costly. But through unsupervised dialogue structure learning, dialogue structure can be automatically obtained, reducing the cost of developers constructing dialogue process. The learned dialogue structure can be used to promote the dialogue generation of the downstream task system, and improve the logic and consistency of the dialogue robot's reply.In this paper, we propose a Bert-based unsupervised dialogue structure learning algorithm DSBERT (Dialogue Structure BERT). Different from the previous SOTA models VRNN and SVRNN, we combine BERT and AutoEncoder, which can effectively combine context information. In order to better prevent the model from falling into the local optimal solution and make the dialogue state distribution more uniform and reasonable, we also propose three balanced loss functions that can be used for dialogue structure learning. Experimental results show that DSBERT can generate a dialogue structure closer to the real structure, can distinguish sentences with different semantics and map them to different hidden states.

\end{abstract}


\section{Introduction}
\label{Sec:Introduction}

\begin{figure}[htbp]
\centering
\begin{minipage}[t]{0.9\linewidth}
\centering
\includegraphics[width=\linewidth]{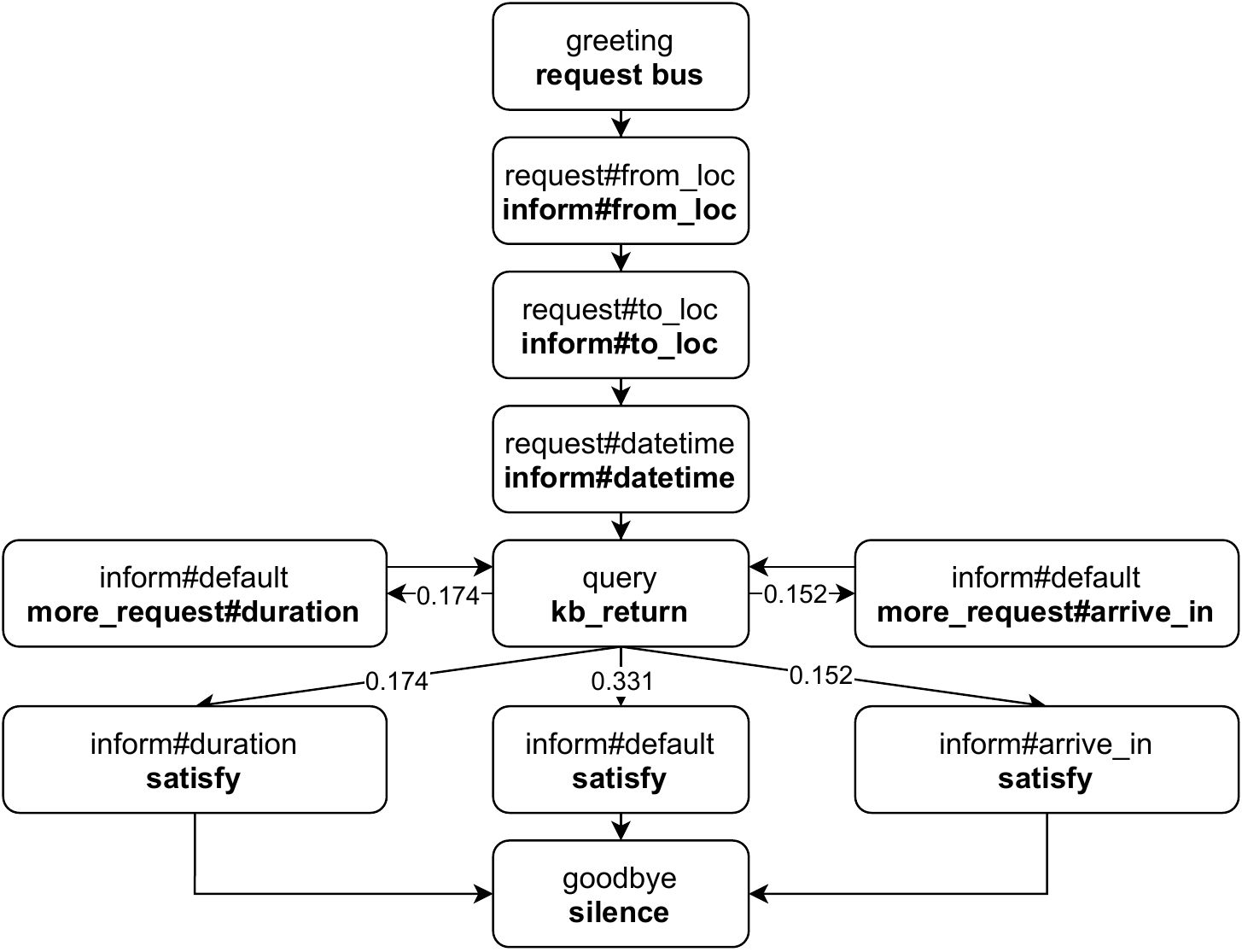}
\subcaption{Bus information request.}
\end{minipage}

\begin{minipage}[t]{0.9\linewidth}
\centering
\includegraphics[width=\linewidth]{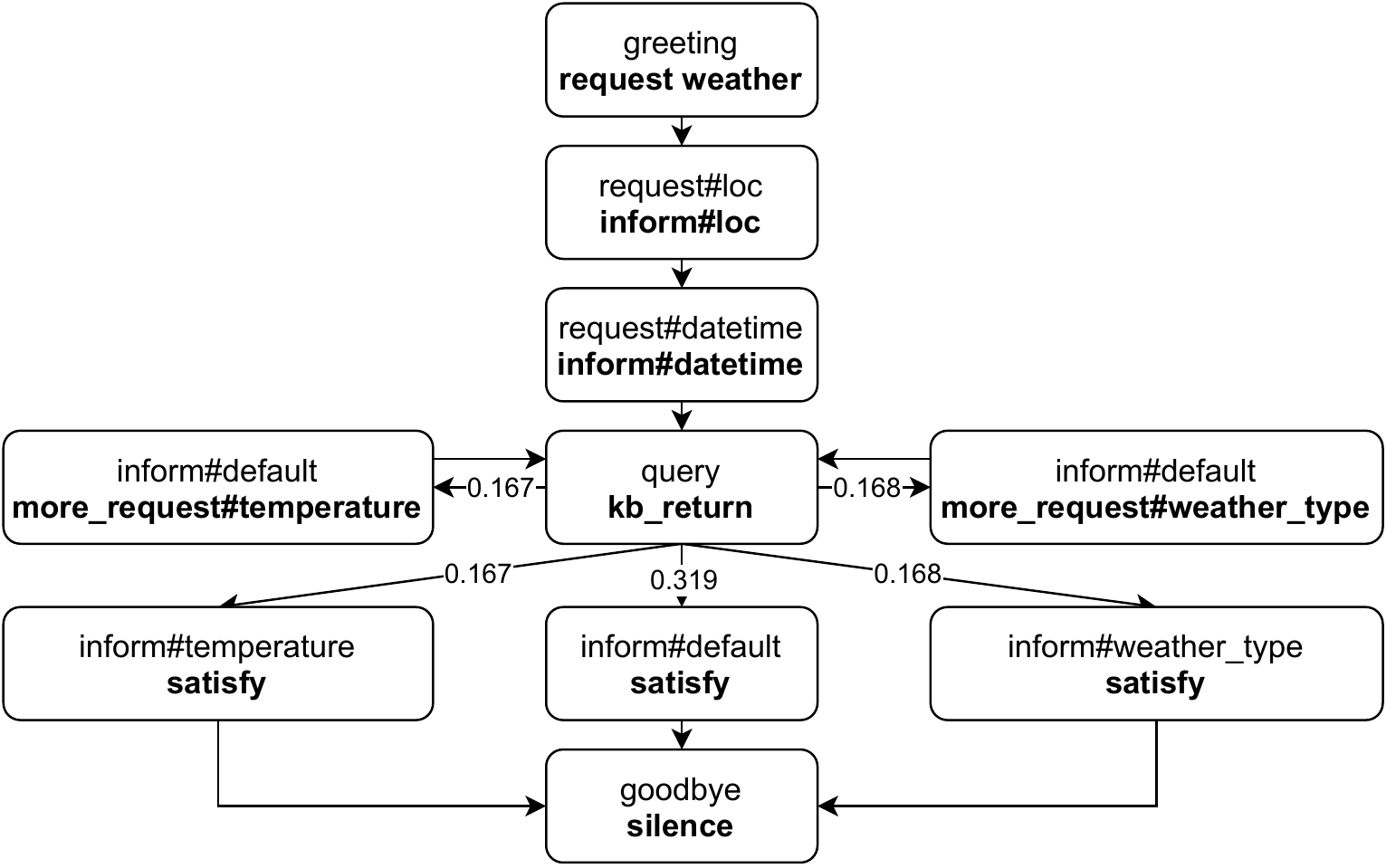}
\subcaption{Weather information request.}
\end{minipage}
\caption{Original dialogue structure of SimDial dataset \cite{2018Zero} and the user's actions are  marked in bold.Figure (a) is the bus information request data and Figure (b) is the weather information request data. \textit{query} and \textit{kb\_return} indicate the system API call and the return result in dialog.}
\label{Figure 1}
\end{figure}

\begin{figure}[htbp]
\centering
\includegraphics[width=\linewidth]{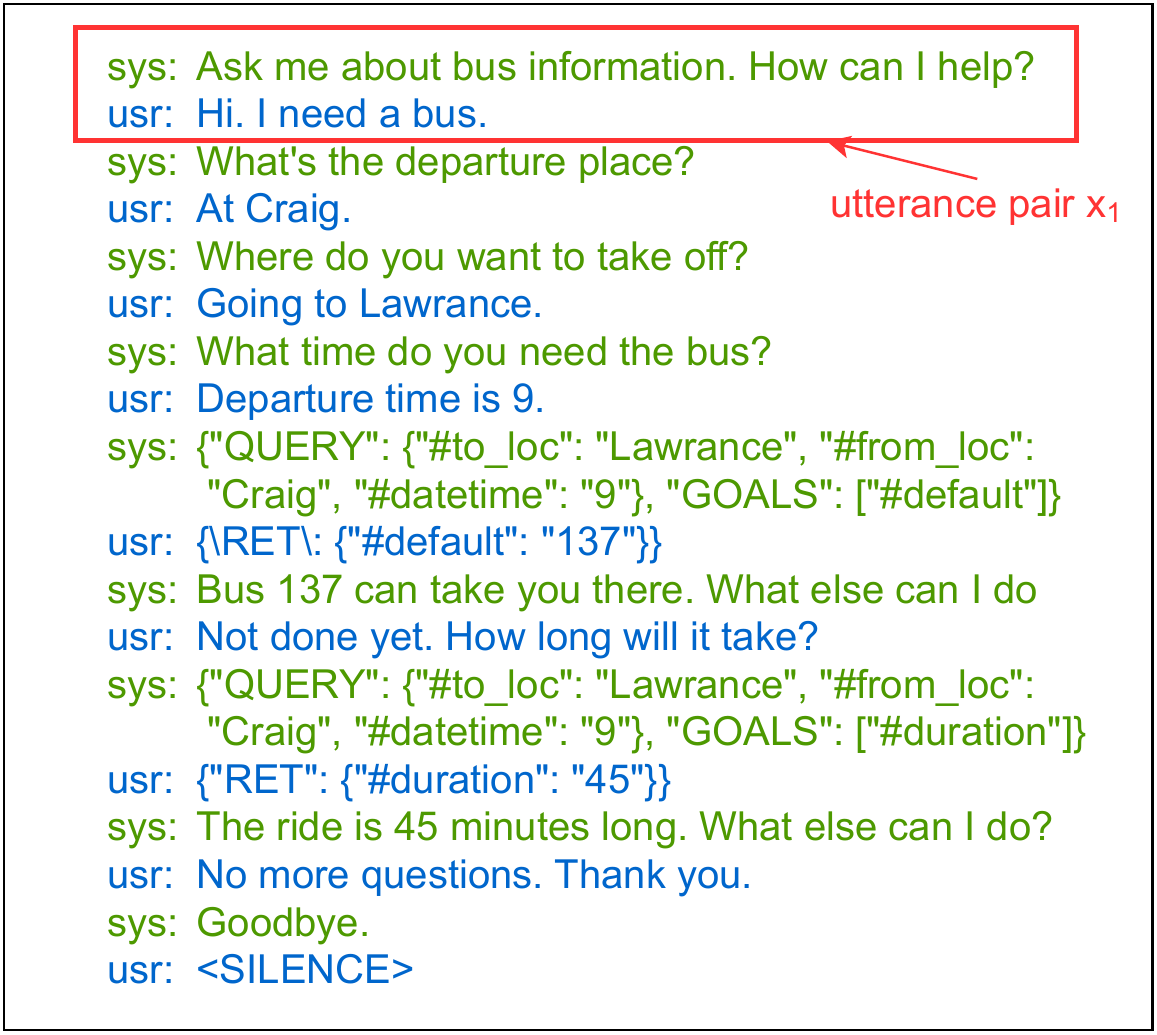}
\caption{A dialog from the bus information request dataset in SimDial. \textit{QUERY} and \textit{RET} are the system API call and result return respectively}.
\label{Fig:EXAMPLE}
\end{figure}

\IEEEPARstart{D}ialogue structure is very helpful for downstream tasks such as dialogue system design \cite{2007Using}, dialogue analysis \cite{grosz1986attention} and dialogue summary \cite{murray2005extractive} \cite{liu2010dialogue}. At present, the dialog structure in dialog system often requires language experts to manually design based on relevant professional knowledge, which takes a lot of time and manpower. Automatic learning dialogue structure by unsupervised learning can reduce the cost of design systems and support a wide variety of dialogue. The dialog structure is typically composed of a probability transfer matrix of latent states, and each utterance pairs of dialogue (including user utterances and system utterances) belongs to a latent state, which has an impact on future status sequences and conversations. Fig \ref{Figure 1} shows the Original dialogue structure of dialogue from SimDial dataset\cite{zhao2018zero} and Fig \ref{Fig:EXAMPLE} shows a dialogue example in the SimDial dataset.

A large number of previous research has studied unsupervised methods to model the latent dialog structure. Early study based on hidden markov model (HMM) algorithm \cite{chotimongkol2008learning} \cite{ritter2010unsupervised} \cite{zhai2014discovering}, usually combined HMM with language model and topic model. Such algorithms can capture the dependencies in the conversation through HMM, but lack the ability to discriminate semantic complex sentences. The model based on variational recurrent neural network is the most commonly used unsupervised dialogue structure learning algorithm, such as VRNN \cite{shi2019unsupervised}, SVRNN \cite{qiu2020structured}. VRNN uses VAE as a cell of recurrent neural network (RNN) and each VAE cell includes encoder and decoder. VRNN maps each utterance pair into latent state via encoder and restores the original sentence according to the latent state by decoder. SVRNN incorporates structured attention layers into VRNN network to enable the model to focus on different portions of the conversation, resulting in a more reasonable dialogue structure than VRNN.

Although VRNN and SVRNN have a good ability to learn dialogue structure, there are still some problems. First, the loss function is not very targeted. The loss function of VRNN and SVRNN includes three parts,  KL-divergence loss between transition probabilities, loss function of language model and loss function of bag-of-word model. The kl divergence loss function measures the difference between the  prior state distribution and the predicted state distribution, but both of these probabilities are generated by the model, which has weak constraints on the model. When training language model, the latent state is used as the initial state of RNN, and then next word is predicted based on previous word. This process model can easily predict the next word directly from previous word, which weakens the ability to learn latent state. Second, VRNN and SVRNN lack necessary constraints and are easy to fall into local optimum. During our experiment, we found that VRNN and SVRNN sometimes fall into local optimum. Many utterance with completely different semantics are assigned to same latent state. In serious cases, all utterance are only assigned to 2~3 latent states.

In this paper, we propose DSBERT (Dialogue Structure BERT) model, using pre-trained language model BERT \cite{devlin2018bert} as the Encoder and Decoder. Bert has strong language representation and feature extraction ability, which can make our model learn semantic information effectively and make our model structure more concise and clear. In order that the latent state generated by DSBERT is highly discriminative and contains sufficient semantic information, we directly use the latent state to reconstruct each round of utterance. At the same time, in order to prevent the unsupervised algorithm from falling into local optimum and to enhance the robustness of the model, we propose three balanced loss functions, so that the utterance can be reasonably distributed in different latent states. We also use TF-IDF to extract the keywords of each utterance and concat the keywords with utterance to make the model more stable during the initial training.

Our contributions include: (1) We propose a BERT-based unsupervised dialogue structure learning framework. The model structure is more concise and effective than VRNN and can better integrate contextual information to learn dialogue semantics. (2) We propose three balanced loss functions that can be used for dialogue structure learning, which can effectively prevent the model from falling into local optimal solution and make utterances reasonably distributed in different states. (3) We propose a scheme of adding keywords to make the preliminary process of unsupervised dialogue structure learning more stable.

\section{Related Work}
\label{Sec:Relate Work}

In the early research work of dialogue structure learning, a common way is to learn dialogue structure based on artificially labeled data \cite{jurafsky1997switchboard}. On the one hand, this method requires a lot of person and time to annotate data. On the other hand, the requirements of users are very diverse and manually labeled data cannot satisfy all users. Therefore, it is meaningful to study unsupervised dialogue structure learning and automatically extract semantic structure from massive unsupervised dialogue data. Most of the early researches on unsupervised dialogue structure learning were based on hidden markov model (HMM). Ritter et al. Proposed the first unsupervised dialogue action learning method for open domain data \cite{ritter2010unsupervised}. The model adds additional word information to the interactive data and combines the conversation model and topic model on HMM. Zhai and Williams proposed three unsupervised methods for learning the dialogue structure of task-oriented dialogue \cite{zhai2014discovering}. The HMM model is used to model hidden state and the topic model is used to connect words and states. However, these models are based on simple HMM and lack the ability to model highly non-linear, while highly non-linear neural network models can capture more semantic information and more dynamics in dialogue.

\begin{figure*}[ht]
\centering
\includegraphics[scale=0.8]{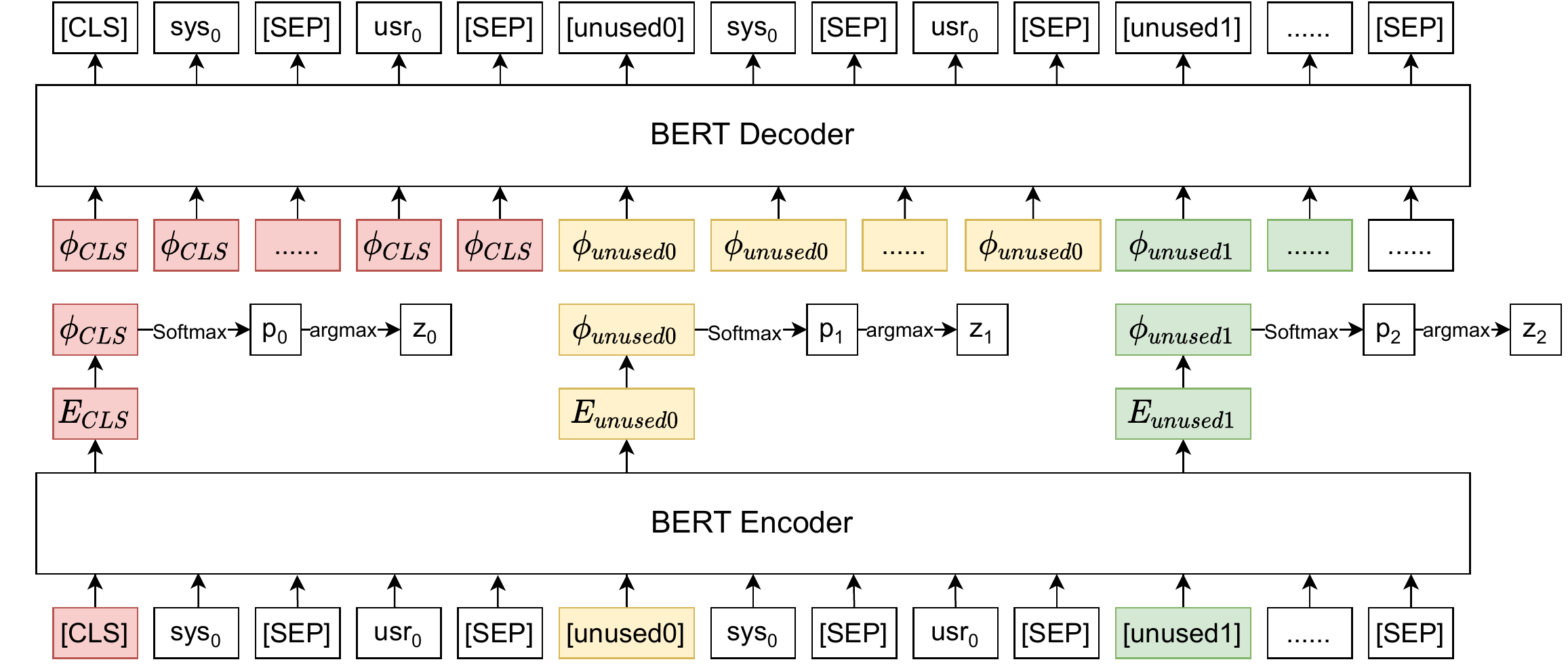}
\caption{The structure of DSBERT}
\label{Fig:DSBERT}
\end{figure*}

Nowadays, in the field of unsupervised dialogue structure learning, neural network-based models are more advanced. Variational autoencoders (VAE)\cite{kingma2013auto} \cite{2016Tutorial} \cite{kingma2014semi} is widely used in various natural language processing tasks because of its interpretable model structure and good generation ability. Chung et al. combined VAE with RNN and proposed a variational recurrent neural network (VRNN) \cite{chung2015recurrent}. VRNN can connect neural networks with traditional bayesian methods and has a nonlinear output at every moment. Shi et al. proposed using a modified VRNN model with discrete latent vectors to extract the semantic structure of task-oriented dialogue \cite{shi2019unsupervised}. Assuming that the dialogue structure is composed of a series of latent states. Each utterance of dialogue belongs to a latent state, and latent state will affect state and content of the utterance at the next moment. Qiu et al. combined structured attention on VRNN and proposed SVRNN \cite{qiu2020structured}. Structured attention enables model to focus on utterance representations at different moments. SVRNN can learn not only semantic structures but also interactive structures by choosing appropriate structural biases or constraints.

Recently, models based on transformer \cite{vaswani2017attention} have achieved state-of-the-art (SOTA) results in various natural language processing tasks. Unlike RNN, transformer processes context information in parallel through the self-attention structure. The pre-trained language model BERT \cite{devlin2018bert} is based on the transformer's encoder, and the model is pre-trained through two self-supervised training tasks, Masked LM and Next Sentence Prediction (NSP). Then the pre-trained model can be fine-tuned in downstream tasks and achieve impressive results. In order to allow BERT to handle longer sequences, many BERT variants for long sequences have also been proposed, such as Longformer \cite{beltagy2020longformer}, Big Bird \cite{zaheer2020big}. These methods use the Sparse attention mechanism to reduce the time complexity of self-attention to linear.

The DSBERT framework we propose is based on BERT model. A BERT encoder is used to convert the dialogue into latent states, and then a BERT decoder is used to restore the entire dialogue according to latent states. DSBERT can be combined with different BERT variant models. For example, DSBERT can use Longformer as Encoder and Decoder when facing long text. We compared DSBERT and other unsupervised dialogue structure extraction algorithms and found that DSBERT can generate a more reasonable dialogue structure and is more stable. It shows that DSBERT has a strong ability to learn the dependency relationship between utterances, and has a strong ability to distinguish different semantic sentences.

\section{DSBERT}
\label{Sec:DSBERT}

The structure of DSBERT is shown in Fig \ref{Fig:DSBERT}. It is an Encoder-Decoder structure and consists of two BERT models. Encoder predicts the latent state of each utterance in dialogue, and then decoder uses the latent states to restore the original dialogue. Through DSBERT, the state sequence $z_1,z_2,...,z_n$ of the dialog can be obtained, and the state transition probability matrix can be estimated through the state sequences. We will describe more details about the key components of our model in the following subsections.

\subsection{Problem Definition}

Given a corpus $D={\{X_1,X_2,...,X_{|D|}\}}$ that containing $|D|$ task-oriented dialogues, each dialogue $X=[x_1,x_2,...,x_t]$ is composed of a sequence of utterance pairs, the utterance pair at moment $i$ includes system utterance $s_i$ and user utterance $u_i$, $x_i=[s_i,u_i]$. Fig \ref{Fig:EXAMPLE} shows a dialog from the bus information request dataset in SimDial \cite{2018Zero}. Our goal is to predict the latent state $z_i$ for each utterance pair in the dialog. and then based on the status of all dialogs The state sequence $[z_1,z_2,...,z_t]$ constructs the flow of the entire task-based dialogue. We can also construct the entire task-oriented dialogue structure according to the state sequence of all dialogues.

\subsection{Encoder}

Suppose we have a dialogue $X=[x_1,x_2,...,x_t]$, $x_i=[s_i,u_i]$, DSBERT needs to predict latent state of each utterance pair and obtain the state sequence $Z=[z_1,z_2,...,z_t]$ of the entire dialogue $Z=[z_1,z_2,...,z_t]$. DSBERT uses BERT as encoder to obtain latent state distribution. We concatenate all utterances of the dialog session in order, and use the concatenated token sequence as the input of BERT. There is a special token used to predict latent state before each utterance pair. For the first utterance special token is CLS, and the remaining utterances use unused0, unused1,... as special tokens. As shown in the Equation \ref{equ:input}.

\begin{equation}
\begin{split}
\label{equ:input}
    input = [\rm{CLS}, s_1, u_1, \rm{SEP}, \rm{unused0}, s_2, u_2, \\
    \rm{SEP},..., \rm{unused(t-1)}, s_t, u_t, \rm{SEP}]
\end{split}
\end{equation}

BERT can get the embedding matrix $E \in R^{L \times d}$ of the input sequence, where $L$ is the total length of the input sequence and $d$ is the BERT output feature dimension. Then we use the fully connected neural network to project the embedding matrix $E$ into the latent state space to obtain the feature matrix $\Phi \in R^{L \times n_{state}}$. We take out the feature $[\phi_0, \phi_1, ..., \phi_{t-1}]$ corresponding to $[\rm{CLS,unused0,unused1,...,unused(t-2)}]$ from $\Phi$, and get the probability distribution $[p_0, p_1, ..., p_{t-1}]$ for each utterance pair by the softmax function. Therefore, the state $z_i$ corresponding to utterance pair $x_i$ is the state with the highest probability in $p_i$, as shown in the Equation \ref{equ:z_i}.

\begin{equation}
\label{equ:z_i}
\begin{split}
    \Phi &= Feed-Forward(E) \\
    p_i &= softmax(\phi_i)\\
    z_i &= \mathop{\rm{argmax}}\limits_{n}[p_i(n)]
\end{split}
\end{equation}

\subsection{Decoder}

In the previous section, we obtained the feature vectors $\phi_0, \phi_1, ..., \phi_{t-1}$ of special token before each utterance. To ensure that these feature vectors can retain the information of dialogue, we use another BERT model acts as decoder to restore the entire dialogue based from $\phi_0, \phi_1, ..., \phi_{t-1}$. The feature vector $\phi_i$ corresponding to utterance $x_i$ is copied $l_i$ times, and $l_i$ is the number of tokens in utterance $x_i$. After copying, we can get a new feature matrix $\tilde{\Phi} \in R^{ L \times n_{state}}$. Then we apply Gumbel-Softmax \cite{2016Categorical} on $\tilde{\Phi}$ to obtain discrete latent state matrix $\tilde{P} \in R^{L \times n_{ state}}$. Finally, we use fully connected network to obtain the input embedding matrix $\tilde{E} \in R^{L \times h}$ of decoder and generate the original dialogue through the decoder. $h$ is the dimension of the BERT token embedding. As shown in the Equation \ref{equ:decoder_input}.

\begin{equation}
\label{equ:decoder_input}
\begin{split}
    \tilde{\Phi} &= [\phi_0;...;\phi_0;...;\phi_{t-1};...;\phi_{t-1}]\\
    \tilde{P} &= gumbel-softmax(\tilde{\Phi})\\
    \tilde{E} &= Feed-Forward(\tilde{P})
\end{split}
\end{equation}

The encoder of DSBERT only uses the features of special token to predict the latent states, so the decoder also uses the features corresponding to special token to restore the dialogue information. The encoder of DSBERT only uses the features of special token to predict the latent states. In the decode stage, we also only use the features of special token to restore the dialogue. We only use special token to restore dialogue mainly for the following reasons: First, this can strengthen the ability of the special token to store dialogue information. Second, when training with language model as VRNN, the model can directly predict the next word based on the previous word without using the latent state, which weakens the training of the latent state.

\subsection{Loss}

The loss function of DSBERT is composed of two parts: Mask Language Model Loss and a Balance Loss. Balance Loss is used to ensure the latent state distribution balance. We first introduce Mask Language Model Loss. DSBERT uses the decoder to predict each token of the original dialogue, and then calculates the cross entropy, denoted as $L_{MLM}$, as shown in the equation \ref{equ:loss_mlm}, and $\theta$ means The parameter of decoder, $\tilde{E}$ is the input of Decoder, and $|V|$ represents the dictionary size.

\begin{equation}
\label{equ:loss_mlm}
\begin{split}
    \mathcal{L}_{\text{MLM}} = &- \sum_{i=1}^{L} \log p(m=w_j|\theta,\tilde{E})\\
    &w_j \in [1,2,...,|V|]
\end{split}
\end{equation}

Assuming that the batch of data contains $M$ dialogues, the number of utterance pairs contained in these dialogues is $\{u_0, u_1,..., u_{M-1}\}$, the total number of utterance pairs contained in this batch Is $U=u_0+u_1+...+u_{M-1}$. We can get the probability that utterance pair belongs to a latent state through encoder, the probability matrix is $P \in R^{U \times n_{state}}$. In actual use, a model trained only through the language model loss is easy to fall into local optimal value. For example, all dialogues are distributed in a few latent states, and it is even possible that all dialogues are allocated to one latent state. This phenomenon also occurs in VRNN and SVRNN. We hope that the probability matrix $P$ should be as balanced as possible in the latent state. Therefore, in addition to Mask Language Model Loss, a Balance Loss must be used to avoid the situation where all dialogues are marked as a few latent states at the same time. We propose Three balanced loss functions.

(1) {\bf{Balance \& KL loss}}. We use Balance Regularization function mentioned in \cite{2020Enhanced} to make $P$ more balanced, the equation is $||P||_b = \sum_{j=1}^{n_{state}} {(\sum_{i=1}^{U} p_{ij})}^2$. But we found in experiment that simply using Balance Regularization will easily make $P$ become a uniform distribution, so we also need to combine Balance Regularization and KL divergence.$P$ is the predicted probability distribution, and a target matrix $T \in R^{U \times n_{state}}$ needs to be constructed to calculate the KL divergence. We select the latent state with the highest probability for each row of $P$, then change the position with the highest probability in $T$ to 1, and the remaining positions to 0. Balance \& KL Loss is shown in the equation \ref{equ:loss_balance_kl}.

\begin{equation}
\label{equ:loss_balance_kl}
    \mathcal{L}_{\text{balance\_kl}} = ||P||_b + KL(T||P)
\end{equation}

(2) {\bf{Greedy Balance Loss}}. We use greedy algorithm to construct the target matrix $T_{greedy} \in R^{U \times n_{state}}$ according to the probability distribution matrix $P \in R^{U \times n_{state}}$. For each column $j$ of matrix $P$, find the row $i$ with the highest probability, set $T_{greedy}(i,j)$ to 1, and then remove the corresponding row from $P$. Repeat the above process until each line of $P$ has been processed. \textcolor[rgb]{1,0,0}{Pseudo code.} After obtaining the target matrix $T_{greedy}$, calculate the KL divergence of $T_{greedy}$ and $P$, as shown in equation \ref{equ:loss_greedy_balance}. We only use Greedy Balance Loss in the first few epochs to make the model obtain an ideal initial distribution.

\begin{equation}
\label{equ:loss_greedy_balance}
    \mathcal{L}_{\text{greedy\_balance}} = KL(T_{greedy}||P)
\end{equation}

(3) {\bf{Top Balance Loss}}. For each column of the probability matrix $P$, find the row with highest probability. The number of columns in $P$ is $n_{state}$, so $n_{state}$ row will be found. Then take out these rows to form a new probability matrix $P' \in R^{n_{state} \times n_{state}}$, and construct the corresponding target matrix $T' \in R^{ n_{state} \times n_{state}}$. Finally, calculate the KL divergence of $T'$ and $P'$, as shown in equation \ref{equ:loss_top_balance}.

\begin{equation}
\label{equ:loss_top_balance}
    \mathcal{L}_{\text{top\_balance}} = KL(T'||P')
\end{equation}

The loss function of DSBERT is combination of Mask Language Model Loss and any Balance Loss. As shown in equation \ref{equ:loss}, $\lambda$ is the coefficient that controls the strength of balance loss.

\begin{equation}
\label{equ:loss}
    \mathcal{L} = \mathcal{L}_{\text{MLM}} + \lambda \mathcal{L}_{\text{balance}}
\end{equation}

\begin{algorithm}
\caption{Target matrix $T$ of Balance \& KL loss}
\label{algorithm:1}
\KwData{Probability matrix $P \in R^{U \times n_{state}}$}
\KwResult{Target matrix $T \in R^{U \times n_{state}}$}
$r\leftarrow t$\;
$\Delta B^{\ast}\leftarrow -\infty$\;
\While{$\Delta B\leq \Delta B^{\ast}$ and $r\leq T$}{$Q\leftarrow\arg\max_{Q\geq 0}\Delta B^{Q}_{t,r}(I_{t-1},B_{t-1})$\;
$\Delta B\leftarrow \Delta B^{Q}_{t,r}(I_{t-1},B_{t-1})/(r-t+1)$\;
\If{$\Delta B\geq \Delta B^{\ast}$}{$Q^{\ast}\leftarrow Q$\;
$\Delta B^{\ast}\leftarrow \Delta B$\;}
$r\leftarrow r+1$\;}
\end{algorithm}

\subsection{Extract Keywords}

In the case of unsupervised, sometimes it is not easy for the model to distinguish sentences with different semantics. In order to allow the model to better capture the differences of each utterance at the beginning of training, we propose to extract $k$ keywords from utterance and concatenate them in front of the utterance. We use TF-IDF to extract keywords, treat each utterance as a corpus, calculate the corresponding TF-IDF vector, and then take the $k$ words with the highest scores in TF-IDF vector as keywords.

\section{Experiments}

\subsection{Datasets}

We use SimDial English task-oriented dialogue dataset \cite{2018Zero} and RiSAWOZ Chinese task-oriented dialogue dataset \cite{quan-etal-2020-risawoz} to verify the performance of DSBERT. The SimDial dataset contains four task: bus, weather, movie, restaurant. For each utterance pair in SimDial data (including system utterance and user utterance), we use the system action and user action provided by the dataset as true labels of the corresponding utterance pair, as shown in Figure \ref{Fig:EXAMPLE}.

the table \ref{table:SimDial} shows the statistical information of the SimDial data set. We also manually annotated the RiSAWOZ Chinese task-based dialogue dataset. The statistics of the RiSAWOZ dataset are shown in the table \ref {table:RiSAWOZ}. Table \ref{table:SimDial} shows the information of SimDial dataset. We also manually annotated the RiSAWOZ Chinese task-oriented dialogue dataset. The information of the RiSAWOZ dataset is shown in table \ref{table:RiSAWOZ}.

\begin{table*}[ht]
    \caption{SimDial Dataset}
    \centering
    \resizebox{\textwidth}{15mm}{
    \begin{tabular}{cccccccc}
    \toprule
    Data & Turns(min) & Turns(max) & Turns(mean) & Tokens(min) & Tokens(max) & Tokens(mean) & Labels \\
    \midrule
    Bus  & 7 & 13 & 8.04 & 59 & 150 & 87.96 & 14 \\
    Weather  & 6 & 10 & 7.05 & 45 & 131 & 75.46 & 12 \\
    Movie  & 7 & 13 & 9.01 & 57 & 156 & 95.24 & 22 \\
    Restaurant  & 6 & 12 & 7.99 & 48 & 146 & 89.67 & 20 \\
    \bottomrule
    \end{tabular}
    }
    \label{table:SimDial}
\end{table*}

\begin{table*}[ht]
    \caption{RiSAWOZ Dataset}
    \centering
    \resizebox{\textwidth}{15mm}{
    \begin{tabular}{cccccccc}
    \toprule
    Data & Turns(min) & Turns(max) & Turns(mean) & Tokens(min) & Tokens(max) & Tokens(mean) & Labels \\
    \midrule
    Tourist attractions & 3 & 11 & 5.3 & 77 & 437 & 187.53 & - \\
    hospitals  & 4 & 11 & 5.79 & 101 & 474 & 212.71 & - \\
    computers  & 4 & 13 & 7.87 & 136 & 508 & 256.32 & - \\
    counseling classes  & 4 & 11 & 6.29 & 116 & 448 & 220.22 & - \\
    \bottomrule
    \end{tabular}
    }
    \label{table:RiSAWOZ}
\end{table*}

\subsection{Baselines}

We choose the following five unsupervised dialogue structure learning methods as Baselines.

(1) {\bf{HMM}}. HMM is a traditional algorithm for modeling hidden states and uses topic models to connect words and states.

(2) {\bf{K-Means}}. We use K-Means to cluster utterance pairs, and the cluster id obtained is the latent state of corresponding utterance pair.

(3) {\bf{D-VRNN}} \cite{shi2019unsupervised}. VRNN obtains other utterance information through the RNN structure and maps the utterance to latent state through VAE.

(4) {\bf{DD-VRNN}} \cite{shi2019unsupervised}. DD-VRNN is similar to D-VRNN, but the method to calculate the state transition prior is different.

(5) {\bf{SVRNN}} \cite{qiu2020structured}. SVRNN is the current SOTA for unsupervised task-oriented dialogue structure learning. It adds a structured attention mechanism to VRNN.

\subsection{Evaluation Metrics}

We use the two evaluation metrics\cite{qiu2020structured}: Structure Euclidean Distance (SED) and Structure Cross-Entropy (SCE) to measure the performance of the model. It should be noted that these two evaluation metrics require the true dialogue structure labels of data. For SimDial and RiSAWOZ, we have already marked their true labels.

Suppose there are $M$ latent states learned by the model, which are $\{s'_i, i=1,2,..., M\}$, and there are N true latent states from the dataset, which are $\{s_i, i=1,2,..., N\}$. The state sequence of all dialogues can be learned through the model, and then the predicted state transition probability matrix $T' \in R^{M \times M}$ can be calculated by counting the occurrence probability of bi-grams in the state sequence, where $T'_{ ij} = \frac{(s’_i,s'_j)}{(s'_i)} $. According to the true state sequence, the real state transition probability matrix $T \in R^{N \times N}$ can also be calculated.

Then we need to calculate the mapping probability matrix $P_{s_i, s'_i} \in R^{N \times M}$ between the real state and the predicted state. It can be calculated through dividing the number of utterances that have the ground truth state $s_i$ and learned state $s'i$ by number of utterances with the ground truth state $s_i$. A similar method can also be used to calculate the mapping probability matrix $P_{s'_i, s_i} \in R^{M \times N}$ between the predicted state and the real state.

To calculate SED and SCE, the state transition probability matrix $T'$ needs to be mapped to the real state and obtain the transition probability matrix $T'' \in R^{N \times N}$, see the equation \ref{equ:pred_trans}.

\begin{equation}
\label{equ:pred_trans}
    T''_{s_a,s_b} = \sum_{i,j \in \{1,2,...,M\}} P_{s_a, s'_i} \cdot T'_{s'_i, s'_j} \cdot P'_{s'_j, s_b}
\end{equation}

SED and SCE are calculated by the equation \ref{equ:SED_SCE}:

\begin{equation}
\label{equ:SED_SCE}
\begin{split}
    SED &= \frac{1}{N} \sqrt{\sum_{a,b \in \{1,2,...,N\}} (T''_{s_a,s_b}-T_{s_a,s_b})^2} \\
    SCE &= \frac{1}{N} \sum_{a,b \in \{1,2,...,N\}} - \text{log}(T''_{s_a,s_b})T_{s_a,s_b}
\end{split}
\end{equation}

\subsection{Extracted Structure}

Extracted structure.

\subsection{Result}

We compare the performance of DSBERT and various baselines on the SimDial English dataset and RiSAWOZ Chinese dataset. The experimental results of SimDial are shown in Table \ref{table:SimDial_result}.

\begin{table*}[ht]
    \caption{Experimental results of SimDial}
    \centering
    \begin{tabular}{|c|c|c|c|c|c|c|c|c|}
    \hline
    \multirow{2}{*}{Model} & \multicolumn{4}{c|}{SED} & \multicolumn{4}{c|}{SCE} \\
    \cline{2-9}
    & bus & weather & movie & restaurant & bus & weather & movie & restaurant \\
    \hline
    HMM & - & - & - & - & - & - & - & - \\
    \hline
    KMeans & - & - & - & - & - & - & - & - \\
    \hline
    VRNN & 0.217 & 0.233 & 0.165 & 0.166 & 1.661 & 1.606 & 1.488 & 1.391 \\
    \hline
    SVRNN & 0.213 & 0.228 & 0.162 & 0.176 & 1.600 & 1.548 & 1.449 & 1.529 \\
    \hline
    DSBERT+Balance KL & {\bf 0.160} & {\bf 0.133} & {\bf 0.133} & {\bf 0.116} & {\bf 0.583} & {\bf 0.430} & {\bf 0.628} & {\bf 0.526} \\
    \hline
    DSBERT+Greedy Balance & 0.198 & 0.218 & 0.164 & 0.139 & 1.349 & 1.358 & 1.448 & 1.005 \\
    \hline
    DSBERT+Top Balance & 0.199 & 0.196 & 0.163 & 0.152 & 1.326 & 1.150 & 1.391 & 1.079 \\
    \hline
    \end{tabular}
    \label{table:SimDial_result}
\end{table*}

\section{Conclusion}
\label{sec:conclusion}
In this paper, we analyze the performance differences between various pre-trained language models fine-tuned on question answering datasets of varying levels of difficulty. Further, we propose an ensemble model and compare its performance to other models. Experimental results show the effectiveness of our methods and shows that RoBERTa and an auxiliary BiLSTM layer both improve model performance in question answering. We see the highest F1-score on RoBERTa and BART model across all datasets. We also observe at least a $1\%$ increase in F1-score over the BERT base model when a BiLSTM layer is added on. Future work includes extending our model to incorporating additional attention mechanisms and potentially utilizing the MatchLSTM architecture to create a better performing ensemble model.

\section{Acknowledgements}
\label{sec:acknowledgements}
This work was supported in part by the Department of Defense and the Army Educational Outreach Program.

\bibliographystyle{IEEEtran}
\bibliography{Bib}

\end{document}